\documentclass[runningheads,a4paper]{llncs}

\usepackage{times}
\usepackage{amssymb}
\usepackage{graphicx}
\usepackage{url}

\usepackage{array}
\usepackage{booktabs}
\usepackage{multirow}
\usepackage{makecell}
\usepackage{tabularx}
\usepackage{svg}
\usepackage[inline]{enumitem}
\usepackage{todonotes}
\usepackage{float}
\usepackage{hyperref}
\usepackage[british]{babel}
\hypersetup{plainpages=false,
            pageanchor=true,
            colorlinks=true,
            linkcolor=blue,
            citecolor=blue,
            anchorcolor=blue,
            menucolor=blue,
            filecolor=blue,
            urlcolor=blue,
            linktocpage=true,
            breaklinks=true,
            unicode=true}
\usepackage[htt]{hyphenat}

\newcommand{\spacy}{spaCy}
\newcommand{\huspacy}{HuSpaCy}
\newcommand{\emtsv}{\texttt{emtsv}}
\newcommand{\magyarlanc}{\texttt{magyarlanc}}
\newcommand{\udpipe}{UDPipe}
\newcommand{\stanza}{Stanza}
\newcommand{\trankit}{Trankit}
\newcommand{\udify}{UDify}

\newcommand{\universaldependencies}{Universal Dependencies}
\newcommand{\ud}{UD}
\newcommand{\udhu}{UD-Hungarian}
\newcommand{\szegedcorpus}{Szeged Corpus}
\newcommand{\nerkor}{NYTK-NerKor}
\newcommand{\szegedner}{Szeged NER}

\newcommand{\floret}{\texttt{floret}}
\newcommand{\fasttext}{\texttt{fastText}}
\newcommand{\wordtovec}{\texttt{word2vec}}
\newcommand{\biaffine}{Biaffine}
\newcommand{\lemming}{Lemming}

\newcommand{\fone}{$F_1$}

\newcommand{\hubert}{\texttt{huBERT}}
\newcommand{\embert}{\texttt{emBERT}}
\newcommand{\bert}{\texttt{BERT}}
\newcommand{\xlmroberta}{\texttt{XLM-RoBERTa-large}}

\newcommand{\huspacylg}{\texttt{lg}}
\newcommand{\huspacymd}{\texttt{md}}
\newcommand{\huspacytrf}{\texttt{trf}}
\newcommand{\huspacytrfxl}{\texttt{trf\_xl}}

\newcommand{\keywords}[1]{\par\addvspace\baselineskip
\noindent\keywordname\enspace\ignorespaces#1}

\urldef{\mailsa}\path|gyorgy@orosz.link|
\urldef{\mailsb}\path|{berkecz, gszabo, szantozs, rfarkas}@inf.u-szeged.hu|

\begin{document}

\title{Advancing Hungarian Text Processing with \huspacy: Efficient and Accurate NLP Pipelines}

\titlerunning{Advancing Hungarian Text Processing with \huspacy}


\author{Gy\"{o}rgy Orosz$^*$ \and Gerg\H{o} Szab\'{o}$^*$ \and P\'{e}ter Berkecz$^*$ \and \\ Zsolt Sz\'{a}nt\'{o} \and Rich\'{a}rd Farkas}



\authorrunning{Orosz et al.}


\institute{Institute of Informatics, University of Szeged\break
    2. Árpád tér, Szeged, Hungary \\
\email{gyorgy@orosz.link}\break
\email{\{gszabo,berkecz,szantozs,rfarkas\}@inf.u-szeged.hu}\break
}

\index{Orosz, Gy\"{o}rgy}
\index{Szab\'{o}, Gerg\H{o}}
\index{Berkecz, P\'{e}ter}
\index{Sz\'{a}nt\'{o}, Zsolt}
\index{Farkas, Rich\'{a}rd}

\toctitle{} \tocauthor{}

\maketitle

\def\thefootnote{*}\footnotetext{These authors contributed equally to this work.}\def\thefootnote{\arabic{footnote}}

%
%
%
%
\begin{abstract}
This paper presents a set of industrial-grade text processing models for Hungarian that achieve near state-of-the-art performance while balancing resource efficiency and accuracy. Models have been implemented in the \spacy{} framework, extending the \huspacy{} toolkit with several improvements to its architecture. Compared to existing NLP tools for Hungarian, all of our pipelines feature all basic text processing steps including tokenization, sentence-boundary detection, part-of-speech tagging, morphological feature tagging, lemmatization, dependency parsing and named entity recognition with high accuracy and throughput. We thoroughly evaluated the proposed enhancements, compared the pipelines with state-of-the-art tools and demonstrated the competitive performance of the new models in all text preprocessing steps. All experiments are reproducible and the pipelines are freely available under a permissive license.

\end{abstract}


\section{Introduction} \label{sec:introduction}

Academic research in natural language processing has been dominated by end-to-end approaches utilizing pre-trained large neural language models which are fine-tuned for the particular applications. Although these deep learning solutions are highly accurate, there is an important demand for human-readable output in real-world language processing systems. Industrial applications are frequently fully or partially rule-based solutions, as (sufficient) training data for a pure machine learning solution is not available and each and every real-world application has its own requirements. Moreover, rule-based components provide tight control over the behavior of the systems in contrast to other approaches.

In this paper, we present improvements to a Hungarian text preprocessing toolkit that achieve competitive accuracies compared to the state-of-the-art results in each text processing step. An important industrial concern about large language models is the computational cost, which is usually not worth the accuracy gain. Transformer-based language models require far more computational resources than static word vectors, and their running costs are typically orders of magnitude higher. Furthermore, practical NLP solutions using large language models often only outperform more lightweight systems by a small margin.

In this work, we focus on text processing pipelines that are controllable, resource-efficient and accurate. We train new word embeddings for cost-effective text processing applications and we provide four different sized pipelines, including transformer-based language models, which enable a trade-off between the running costs and accuracy for practical applications. To make our pipelines easily controllable, we implement them in the \spacy\footnote{\url{https://spacy.io/}} framework \cite{spacy} by extending \huspacy{} \cite{huspacy:2021} with new models. 

\section{Background} \label{sec:background}
\subsection{Specification for Language Processing Pipelines for Industrial Use} \label{sec:background:specification}

Text processing tools providing representation for hand-crafted rule construction should consist of tokenization, sentence splitting, PoS tagging, lemmatization, dependency parsing, named entity recognition and word embedding representation. These solutions have to be accurate enough for real-world scenarios while they should be resource-efficient at the same time. Last but not least, modern NLP applications are usually multilingual and should quickly transfer to a new language. This can be provided by relying on international annotation standards and by the integration into multilingual toolkits. 

\subsection{Annotated Datasets for Preprocessing Hungarian Texts} \label{sec:background:datasets}

According to Simon et al. \cite{metanet}, Hungarian is considered to be one of the best supported languages for natural language processing. In 2004, the \szegedcorpus{} \cite{szc} was created, comprising 1.2 million manually annotated words for part-of-speech tags, morphological descriptions, and lemmata. Subsequently, these annotations were extended \cite{szeged-treebank} with dependency syntax annotations. In 2017, a small section of the corpus was manually transcribed to be a part of the \universaldependencies{} (\ud) project \cite{nivre-etal-2020-universal}. Around the same time, the entire corpus was automatically converted from the original codeset to the universal part-of-speech and morphological descriptions \cite{vincze-etal-2017-universal}.

\szegedner{} \cite{szarvas-ner}, developed in 2006, was the first Hungarian named entity recognition corpus, consisting of 200,000 words of business and criminal news. In recent years, \nerkor{} \cite{nerkor} extended the possibilities of training and benchmarking entity recognition systems for Hungarian with a one million word multi-domain corpus.

\subsection{Multilingual NLP Toolkits} \label{sec:background:toolkits}

Thanks to the \ud{} project, it is now possible to easily construct multilingual NLP pipelines. Among the most commonly utilized toolkits are \udpipe{} \cite{udpipe}, \stanza{} \cite{stanza}, \udify{} \cite{udify}, \trankit{} \cite{van2021trankit} and \spacy.

On the one hand, these systems exhibit a high degree of algorithmic diversity. They can be classified into two distinct groups based on their utilization of neural networks. \udpipe, \spacy{} and \stanza{} apply older, but faster architectures built on word embeddings employing convolutional and recurrent layers, respectively. On the contrary, \udify{} and \trankit{} leverage transformer-based large language models, with the former using multilingual \bert{} \cite{devlin2018bert} while the latter utilizing \xlmroberta{} \cite{conneau2019unsupervised}.

On the other hand, these frameworks are typically limited by the fact that they rely solely on the \universaldependencies{} datasets, which may present a disadvantage in languages such as Hungarian, which have large corpora incompatible with \ud. Each of the above-mentioned systems shares this limitation, moreover, \spacy{} does not offer a Hungarian model at all, due to the restrictive license of the \udhu{} corpus. Regarding named entity annotations, \stanza{} is the only tool supporting NER for Hungarian.

\subsection{Hungarian Language Processing Tools}  \label{sec:background:hunlp}

The landscape of the Hungarian text processing systems was similar to that of English before the ``industrial NLP revolution''. There were a number of standalone text analysis tools \cite{metanet} capable of performing individual text processing tasks, but they often did not work well with each other.

There were only two Hungarian pipelines that try to serve industrial needs. One of them, \magyarlanc{} \cite{magyaralanc}, was designed for industrial applications offering several desirable features such as software quality, speed, memory efficiency, and customizability. However, despite being used in commercial applications in the real world, it has not been maintained for several years and lacks integration with the Python ecosystem. The other pipeline, called \emtsv{} \cite{emtsv1,emtsv2,emtsv4}, aimed to integrate existing NLP toolkits into a single application, but neither computational efficiency nor developer ergonomics were the main goals of the project. Additionally, while \magyarlanc{} natively uses the universal morphosyntactic features, \emtsv{} can only do this through conversion. Both pipelines use dependency annotation that is incompatible with \universaldependencies, furthermore, none of them can utilize word embeddings or large language models, which have become increasingly important in recent years.

In contrast, the development of \huspacy{} placed emphasis not only on accuracy, but also on software ergonomics, while also adhering to the international standards established by Nivre et al. \cite{nivre-etal-2020-universal}. Moreover, it is built on \spacy, enabling users to access its full functionality with ease. One significant drawback of this tool is the lack of precise annotations for lemmata, entities and dependencies syntax.

To fulfill the industrial requirements of text processing pipelines, this work is built on the \universaldependencies{} annotation schema and our models are implemented in \spacy{} by extending \huspacy’s text processing model. The detailed documentation, intuitive API, high speed and accuracy of these tools make them an optimal choice for building high-performing NLP models. Additionally, \huspacy{} utilizes non UD compatible corpora as well, which allows for a comprehensive analysis of Hungarian texts.

\section{Methods} \label{sec:methods}
\subsection{\huspacy\textquotesingle s Internals} \label{sec:methods:internals}

\begin{figure}[h]
\centering
\includegraphics[width=\textwidth]{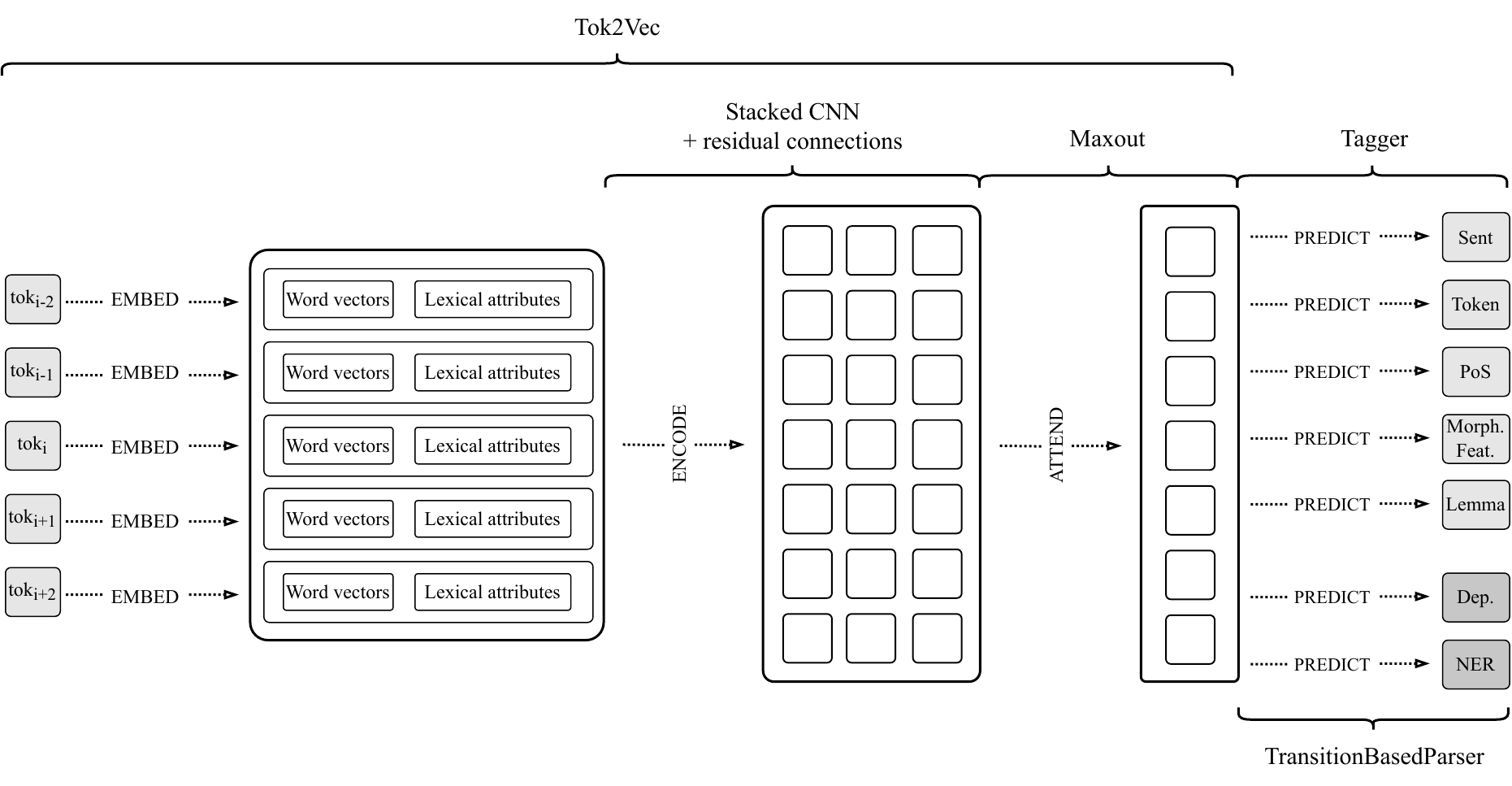}
\caption{The ``embed, encode, attend, predict'' architecture of \spacy}
\label{fig:architecture}
\end{figure}

\huspacy’s main strength lies in the clever usage of available Hungarian linguistic resources and its multi-task learning capabilities inherited from \spacy. Its machine learning approach can be summarized as ``embed, encode, attend, predict''
shown in Figure \ref{fig:architecture} and detailed by \cite{spacy-parser,huspacy:2021}. Tokens are first embedded through the combination of lexical attributes and word vectors, then context encoding is performed by stacked CNN \cite{cnn} layers\footnote{These steps are usually referred to as the Tok2Vec layers.}. Finally, task specific layers are used parallelly in a multi-task learning setup.

Orosz et al. \cite{huspacy:2021} used a three step approach for fully utilizing annotated Hungarian datasets. First, they pre-train the tagger, the lemmatizer and the sentence boundary detection components on a silver standard \ud{} annotated corpus (cf. \cite{vincze-etal-2017-universal}). Then, the Tok2Vec layers of this model are reused by both the NER and the parsing components: the dependency parser and the morphosyntactic taggers are fine-tuned on the \udhu{} dataset, the lemmatizer is trained on the entire \szegedcorpus, while the entity recognizer is further trained on the combination of the \nerkor{} and the \szegedner{} datasets.

\subsection{Improving on the Underlying Language Models} \label{sec:methods:underlying}

\huspacy\textquotesingle s model is built on \wordtovec{} \cite{word2vec} word embeddings, which are known to have limitations in providing meaningful representations for out-of-vocabulary words. This is particularly problematic for morphology-related tasks in agglutinative languages. To enhance this simple approach, a more fine-grained method that uses sub-word embeddings can be employed. \fasttext{} \cite{bojanowski2017enriching} is a widely-used extension of \wordtovec{} that learns sub-token embeddings. In this study, we utilized \floret\footnote{\url{https://explosion.ai/blog/floret-vectors}} which is a \spacy-compatible fork of \fasttext. To train new word vectors, we used the Hungarian Webcorpus 2.0 \cite{hubert}. Two sets of word embeddings were constructed: a 100-dimensional and a 300-dimensional one.

In recent years, there has been a growing interest in transformer-based large language models (LLM), as evidenced by their high performance in text processing models (e.g. \cite{hubert,enevoldsen2021dacy}). With the advent of \spacy\textquotesingle s native support for such architectures and the availability of pre-trained language models for Hungarian, it is now possible to train transformer-based NLP pipelines for Hungarian. Our research is based on two widely used LLMs that provide support for Hungarian. One of these is \hubert{} \cite{hubert}, which has a \bert-like architecture and was trained using monolingual data. The other model is \xlmroberta{}, which has a much larger capacity compared to the former model and was trained on multilingual corpora.

\subsection{Pipeline Component Enhancements} \label{sec:methods:enchancements}

In addition to the use of more powerful language models, we propose fundamental changes to the lemmatization and dependency parsing models, as well as minor improvements to the entity recognizer.

\huspacy\textquotesingle s lemmatizer has been replaced by a new edit-tree-based architecture, recently available in the \spacy{} framework\footnote{\url{https://explosion.ai/blog/edit-tree-lemmatizer}}. This new model builds on the foundations laid out by Müller et al. \cite{muller-etal-2015-joint} (called the \lemming{} model), but has minor differences from it. On the one hand, this reimplementation fully utilizes the framework’s multi-task learning capabilities, which means that the lemmatizer is not only co-trained with PoS and morphological tagging, but also with sentence boundary detection. On the other hand, \spacy\textquotesingle s version lacks standard support for morphological lexicons which \lemming{} benefited from.

We have improved this model in two steps. 
\begin{enumerate*}
    \item A simple dictionary learning method is put in place to memorize frequent \textit{(token, tag, lemma)} triplets of the training data which are then used at prediction time to retrieve the roots of words. 
    \item A common weakness of Hungarian lemmatization methods is addressed. Computing the lemmata of sentence-starting tokens can be challenging for non-proper nouns, as their roots are always lowercase. Thus, we force the model to use the true casing of such words. For example, when computing the root of the sentence starting \textit{Ezzel} `with this' token, our method checks its PoS tag (that is ideally PRON) first, so that it can use the lowercase wordform for generating and looking up edit-trees.
\end{enumerate*}

Moving on, the dependency syntax annotation component is replaced with a model that has higher accuracy for many languages. Although \spacy\textquotesingle s built-in transition-based parser \cite{spacy-parser} has high throughput, it falls short on providing accurate predictions. Graph-based architectures are known to have good performance for dependency parsing (e.g. \cite{altintacs2023improving}), making such methods good enhancement candidates. Furthermore, a \spacy-compatible implementation of Dozat and Manning’s model \cite{biaffine} (referred to as the \biaffine{} parser) has recently been made available, thus we could easily utilize it in our experiments.

Finally, the named entity recognizer has been fine-tuned to provide more accurate entity annotations. This was primarily achieved by using beam-search in addition to the transition-based NER module.

\section{Experiments and Results} \label{sec:experiments}

This section presents the results of several experiments that demonstrate the improvements of our changes and show competitive results compared to well-established baselines. We evaluated pipelines developed on datasets used by the creators of \huspacy: the Hungarian part of the \universaldependencies{} corpus\footnote{Experiments are performed at the v2.10 revision.} was utilized to benchmark the sentence boundary detector, the lemmatizer, the PoS and morphological taggers, and the dependency parser, while the entity recognizer is benchmarked on the combination of the \nerkor{} and the \szegedner{} corpora (similar to \cite{huspacy:2021} and \cite{nerkoreval}). To account for the instability of \spacy’s training process we report the maximum result of three independent runs. 

\subsection{Evaluation of Architecture Improvements} \label{sec:experiments:eval}

The lemmatization accuracy of the original model has been greatly improved through a number of steps discussed in Section \ref{sec:methods:enchancements}. As evidenced in Table \ref{table:lemma}, incorporation of the new neural architecture along with sub-word embeddings produced significant improvements. Furthermore, changing the default behavior of the edit-tree lemmatizer by allowing it to evaluate more than one candidate (see the row \texttt{topk=3}) also resulted in a slightly better performance. In addition, the integration of true-casing led to a considerable improvement, and the use of lemma dictionaries also significantly improved lemmatization scores.

\newcolumntype{R}{>{\centering\arraybackslash}X}
\begin{table}[H] 
	\begin{center}
		\caption{Lemmatization accuracy on the \udhu{} test set of different ablation settings. Rows marked with a ``+'' indicate a new feature added on top of the previous ones. \texttt{topk} is a hyperparameter of the lemmatization model controlling the number of edit-trees considered to be evaluated.}
		\begin{tabularx}{5.75cm}{l R}
			\toprule
			                       & \multicolumn{1}{c}{Lemma Accuracy} \\
			\midrule
			\huspacy               & 95.53\%                            \\
			+ Edit-tree lemmatizer & 95.90\%                            \\
			+ \floret{} 300d vectors    & 96.76\%                            \\
			+ \texttt{topk=3}      & 97.01\%                            \\
			+ True-casing          & 97.30\%                            \\
			+ Learned dictionary   & 97.58\%                            \\
			\bottomrule
		\end{tabularx}
		\vspace{1em}
		\label{table:lemma}
	\end{center}
	\vspace{-2em}
\end{table}

Entity recognition tasks often encounter a challenge in the form of a considerable number of out-of-vocabulary tokens, leading to decreased performance. However, the utilization of \floret{} vectors has proven to be effective in addressing this issue, as indicated by the results in Table \ref{table:ner}. Additionally, the use of beam search allowed the model to take prediction history into account, which slightly improved its efficiency.

\newcolumntype{R}{>{\centering\arraybackslash}X}
\begin{table} [H]
	\begin{center}
		\caption{Evaluation of the entity recognition model improvements on the combination of the \szegedner{} and \nerkor{} corpora.  The rows starting with ``+'' signify the inclusion of a new feature in addition to the existing ones.}
		\begin{tabularx}{5cm}{l R}
			\toprule
			                    & \multicolumn{1}{c}{NER \fone-score} \\
			\midrule
			\huspacy            & 83.68                               \\
			+ \floret{} 300d vectors & 85.53                               \\
			+ Beam search       & 85.99                               \\
			\bottomrule
		\end{tabularx}
		\vspace{1em}
		\label{table:ner}
	\end{center}
	\vspace{-2em}
\end{table}

The results in Table \ref{table:parser} indicate that the improved text representations and the new parsing architecture offer substantial improvements over \huspacy\textquotesingle s outcomes. However, it is worth noting that \spacy’s CNN-based base model is not fully compatible with the \biaffine{} parser’s architecture. Therefore, parsing improvements were benchmarked on top of a transformer-based encoder architecture using \hubert. The results show that the use of \floret{} vectors is beneficial to predict morphosyntactic characteristics and dependency relations, while the use of \hubert-based text representations substantially improves performance across all subtasks. Furthermore, the \biaffine{} parser significantly outperforms its transition-based counterpart, as evidenced by its better attachment scores.

\newcolumntype{R}{>{\centering\arraybackslash}X}
\begin{table}[H] 
	\begin{center}
		\caption{Evaluation of text parsing improvements on the \udhu{} test set. ``+'' indicate a new feature added on top of the existing ones.}
		\begin{tabularx}{9cm}{l R R R R}
			\toprule
			                    & \multicolumn{1}{c}{\makecell{PoS \\ Acc.}} & \multicolumn{1}{c}{\makecell{Morph. \\ Acc.}} & \multicolumn{1}{c}{UAS}   & \multicolumn{1}{c}{LAS}   \\
			\midrule
			\huspacy             & 96.58\%  & 93.23\%     & 79.39 & 74.22 \\
			\huspacy{} + \floret{} 300d vectors & 96.55\%  & 93.93\%     & 80.36 & 74.89 \\
			\huspacy{} + \hubert           & 98.10\%  & 96.97\%     & 89.95 & 83.94 \\
			+ \biaffine{} parser  & 98.10\%  & 96.97\%     & 90.31 & 87.23 \\
			\bottomrule
		\end{tabularx}
		\vspace{1em}
		\label{table:parser}
	\end{center}
	\vspace{-2em}
\end{table}

\subsection{Comparison with the State-of-the-Art} \label{sec:experiments:sota} 

In addition to parsing and tagging correctness, resource consumption is an important consideration for industrial NLP applications. Therefore, following the approach of Orosz et al. \cite{huspacy:2021} we conducted a benchmark study to compare both the accuracy and memory usage as well as the throughput of our models with text processing tools available for Hungarian.

\newcolumntype{R}{>{\centering\arraybackslash}X}
\begin{table}[H]
	\begin{center}
		\caption{Text parsing accuracy of the novel pipelines compared to \huspacy, \stanza, \udify, \trankit{} and \emtsv. Results for non-comparable models are shown in italics.}
		\begin{tabularx}{\textwidth}{l R R R R R R R}
			\toprule
			                  & \multicolumn{1}{c}{\makecell{Sent. \\ \fone-score}}    & \multicolumn{1}{c}{\makecell{PoS \\ Acc.}}         & \multicolumn{1}{c}{\makecell{Morph. \\ Acc.}}      & \multicolumn{1}{c}{\makecell{Lemma \\ Acc.}}       & \multicolumn{1}{c}{UAS}                     & \multicolumn{1}{c}{LAS}                     & \multicolumn{1}{c}{\makecell{NER \\ \fone-score}}      \\
			\midrule
			\textit{\emtsv}   & \textit{98.11} & \textit{89.19\%} & \textit{87.95\%} & \textit{96.16\%} & \textit{--}              & \textit{--}              & \textbf{92.99} \\
			\textit{\trankit} & \textit{98.00} & \textit{97.49\%} & \textit{95.23\%} & \textit{94.45\%} & \textit{\textbf{91.31}} & \textit{\textbf{87.78}} & \textit{--}              \\
			\udify            & --              & 96.15\%          & 90.54\%          & 88.70\%          & 88.03                   & 83.92                   & --              \\
			\stanza           & 97.77          & 96.12\%          & 93.58\%          & 94.68\%          & 84.05                   & 78.75                   & 83.75          \\
			\midrule
			\huspacy          & 97.54          & 96.58\%          & 93.23\%          & 95.53\%          & 79.39                   & 74.22                   & 83.68          \\
			\huspacymd        & 97.88          & 96.26\%          & 93.29\%          & 97.38\%          & 79.25                   & 73.99                   & 85.35          \\
			\huspacylg        & 98.33          & 96.91\%          & 93.93\%          & 97.58\%          & 79.75                   & 74.78                   & 85.99          \\
			\huspacytrf       & 99.33          & \textbf{98.10\%} & \textbf{96.97\%} & 98.79\%          & 90.31                   & 87.23                   & 91.35          \\
			\huspacytrfxl     & \textbf{99.67} & 97.79\%          & 96.53\%          & \textbf{98.90\%} & 90.22                   & 86.67                   & 91.84          \\
			\bottomrule
		\end{tabularx}
		\label{table:compare}
	\end{center}
	\vspace{-3em}
\end{table}

\newcolumntype{R}{>{\raggedleft\arraybackslash}X}
\begin{table}[H] 
	\begin{center}
		\caption[Resource usage of the new models and state-of-the-art of text processing tools available for Hungarian.]{Resource usage\protect\footnotemark{} of the new models and state-of-the-art of text processing tools available for Hungarian. Throughput is measured as the average number of processed tokens per second, while memory usage columns records the peak value of each tool.}
		\begin{tabularx}{8.5cm}{l R R R}
  \toprule
        &                         & \multicolumn{1}{l}{Throughput}  & \multicolumn{1}{c}{\multirow{2}{*}{\makecell{Memory Usage\\(GB)}}} \\
        & \multicolumn{1}{r}{CPU} & \multicolumn{1}{r}{GPU}         & \multicolumn{1}{c}{}                                               \\
  \midrule
			\emtsv        & 113            & --             & 3.9               \\
			\trankit      & 434            & 2119           & 3.7               \\
			\udify        & 129            & 475            & 3.2               \\
			\stanza       & 30             & 395            & 5.3               \\
			\midrule
			\huspacy      & 1525           & 6697           & 3.5               \\
			\huspacymd    & 2652           & 3195           & 1.4               \\
			\huspacylg    & 847            & 3128           & 3.2               \\
			\huspacytrf   & 273            & 2605           & 4.8               \\
			\huspacytrfxl & 82             & 2353           & 18.9              \\
			\bottomrule
		\end{tabularx}
		\label{table:speed}
	\end{center}
\end{table}
\footnotetext{All benchmarks are run on the same environment having AMD EPYC 7F72 CPUs and NVIDIA A100 GPUs.}

First of all, an important result of this study is a base model (referred to as \huspacylg), which achieves a good balance between accuracy and resource usage as seen in Tables \ref{table:compare} and \ref{table:speed}. This pipeline is built on top of the 300d \floret{} vectors and incorporates all the enhancements described above, except for the new parser. Evaluation data demonstrates that the \huspacylg{} pipeline consistently outperforms \stanza{} in all tasks except syntactic dependency relation prediction, which can be explained by the superior parsing model of the latter tool. 

We present the results of a medium-sized model (\huspacymd) as well that is a reduced version of the \huspacylg{} pipeline utilizing the smaller (100d) word embeddings. Surprisingly, the \huspacymd{} pipeline delivers performance similar to that of the larger model. Furthermore, the medium-sized model achieves scores comparable to or higher than those of \huspacy, despite requiring half the memory and exhibiting much higher throughput on CPU. 

Transformer-based pipelines using the graph based dependency parser have the highest scores across all language analysis tasks. Remarkably, despite its smaller capacity, the model based on \hubert{} (\huspacytrf) achieves the highest attachment scores for dependency parsing, while the one using \xlmroberta{} (\huspacytrfxl) provides slightly more accurate PoS tags and named entities.

It is important to consider that not all third-party pipelines in Table \ref{table:compare} are directly comparable to our results, due to differences in the versions of the \udhu{} dataset used to train and evaluate their models. To ensure a fair comparison, \stanza{} and \udify{} have been retrained. On the other hand, we obtained the results of \trankit{} from \cite{van2021trankit} since it would be a demanding task to fine-tune this model. Furthermore, the results of \emtsv\textquotesingle s text parsing components \cite{orosz-novak-2013-purepos,novak2014new,novak-etal-2016-new} cannot be deemed reliable either (cf. \cite{huspacy:2021}), since its components use a different train-test split of the \szegedcorpus. However, this tool’s entity recognition module (\embert{} \cite{embert}) was evaluated by Simon et al. \cite{nerkoreval} using the same settings as in our paper, thus we rely on their assessment. Additionally, state-of-the-art results are also shown in Table \ref{table:compare}. With regard to highest dependency parsing scores, the results of the multilingual \trankit{} system are produced by a parsing model similar to that of ours. As for named entity recognition, \embert{} attains the best \fone{} scores by utilizing a Viterbi encoder that eliminates invalid label sequences from the outputs of the underlying model.

Regarding computational requirements, Table \ref{table:speed} presents findings that demonstrate how \floret{} embeddings can effectively decrease the memory usage of models without compromising their accuracy and throughput. However, it is apparent that enhancing pipeline accuracy frequently results in slower processing speed, as can be observed from the \huspacylg, \huspacytrf{} and \huspacytrfxl{} models. Additionally, our tests also showed that most of the readily available NLP pipelines are not adequately optimized to handle large workloads, which is evident from their low throughput values.

\section{Conclusion} \label{sec:conclusion}

This paper has introduced new industrial-grade text processing pipelines for Hungarian and presented a thorough evaluation showing their (close to) state-of-the-art performance. We have shown that new architectures for lemmatization and dependency parsing and the use of improved text representation models significantly improve the accuracy of \huspacy. The presented models have not only demonstrated high performance in all text preprocessing steps, but the resource consumption of three of our models’ (\huspacymd, \huspacylg, \huspacytrf) makes them suitable for solving practical problems. All of our experiments are reproducible and the models are freely available under a permissive license.

For future work, we consider the following areas of improvement.
\begin{enumerate*}
    \item Transformer\-based pipelines are optimized for accuracy but this could limit their usability due to reduced computational efficiency. We would like to investigate optimizing their size to enhance their resource usage. 
    \item We would like to include more silver standard data to further improve the parsing and tagging scores, as the corpus used to train and evaluate text parsing components is limited in size. 
    \item Our models are mostly trained on news\hyp{}related corpora, which makes user-generated text processing a difficult task. In order to address this challenge, we intend to integrate automatic data augmentation into the training process as a solution.
\end{enumerate*}

\subsubsection*{Acknowledgments.} The authors would like to thank Gábor Berend for his valuable suggestions.
\huspacy{} research and development is supported by the European Union project RRF-2.3.1-21-2022-00004 within the framework of the Artificial Intelligence National Laboratory. 

\bibliographystyle{splncs04}
\bibliography{paper}

\end{document}